\documentclass[letterpaper, 10 pt, conference]{ieeeconf}  

\IEEEoverridecommandlockouts                              

\usepackage{algorithm}        
\usepackage{algpseudocode}    
\usepackage{amsfonts}         
\usepackage{amsmath}          
\usepackage{graphicx}         
\usepackage[pdftex]{hyperref} 
\usepackage[utf8]{inputenc}   
\usepackage{multirow}         
\usepackage{paralist}         
\usepackage{subfig}           
\usepackage{units}            
\usepackage{tikz}             
\usetikzlibrary{chains,shapes.geometric,shapes.misc}

\newtheorem{definition}{Definition}

\newcommand{\sect}[1]{Sec.~\ref{sec:#1}}

\newcommand{\fig}[1]{Fig.~\ref{fig:#1}}

\newcommand{\tab}[1]{Table~\ref{tab:#1}}

\newcommand{\alg}[1]{Alg.~\ref{alg:#1}}

\graphicspath{{img/}}
\algnewcommand\algorithmicswitch{\textbf{switch}}
\algnewcommand\algorithmiccase{\textbf{case}}
\algnewcommand\algorithmicdefault{\textbf{default}}
\algdef{SE}[SWITCH]{Switch}{EndSwitch}[1]{\algorithmicswitch\ #1\ \algorithmicdo}{\algorithmicend\ \algorithmicswitch}%
\algdef{SE}[CASE]{Case}{EndCase}[1]{\algorithmiccase\ #1}{\algorithmicend\ \algorithmiccase}%
\algdef{SE}[DEFAULT]{Default}{EndDefault}[1]{\algorithmicdefault\ #1}{\algorithmicend\ \algorithmicdefault}
\algtext*{EndCase}%
\algtext*{EndDefault}%

\title{\LARGE \bf
  Decentralized Connectivity-Preserving \\
  Deployment of Large-Scale Robot Swarms
}

\author{%
  Nathalie Majcherczyk$^{1}$,
  Adhavan Jayabalan$^{1}$,
  Giovanni Beltrame$^{2}$ and
  Carlo Pinciroli$^{1}$
\thanks{$^{1}$ N. Majcherczyk, A. Jayabalan, and C. Pinciroli are with Robotics Engineering, Worcester Polytechnic Institute, 85 Prescott St., Worcester MA 10609, USA. E-mail: {\tt\small \{nmajcherczyk,} {\tt\small ajayabalan,cpinciroli\}@wpi.edu}}%
\thanks{$^{2}$ G. Beltrame is with Department of Computer and Software Engineering, \'{E}cole Polytechnique Montr\'{e}al, 2900 Edouard Montpetit Blvd, Montr\'{e}al, QC H3T 1J4, Canada. E-mail: {\tt\small giovanni.beltrame@polymtl.ca}}%
}

\begin{document}

\maketitle
\thispagestyle{empty}
\pagestyle{empty}

\begin{abstract}
  We present a decentralized and scalable approach for deployment of a
  robot swarm. Our approach tackles scenarios in which the swarm must
  reach multiple spatially distributed targets, and enforce the
  constraint that the robot network cannot be split. The basic idea
  behind our work is to construct a logical tree topology over the
  physical network formed by the robots. The logical tree acts as a
  \emph{backbone} used by robots to enforce connectivity
  constraints. We study and compare two algorithms to form the logical
  tree: \emph{outwards} and \emph{inwards}. These algorithms differ in
  the order in which the robots join the tree: the outwards algorithm
  starts at the tree root and grows towards the targets, while the
  inwards algorithm proceeds in the opposite manner. Both algorithms
  perform periodic reconfiguration, to prevent suboptimal topologies
  from halting the growth of the tree. Our contributions are
  \begin{inparaenum}[(i)]
  \item The formulation of the two algorithms;
  \item A comparison of the algorithms in extensive physics-based
    simulations;
  \item A validation of our findings through real-robot experiments.
  \end{inparaenum}
\end{abstract}


\section{Introduction}
\label{sec:introduction}

Swarm robotics~\cite{Brambilla2013} is a branch of collective robotics
that studies decentralized solutions for the problem of coordinating
large teams of robots. Robot swarms are a promising technology for
large-scale scenarios, in which performing spatially distributed tasks
would entail prohibitive costs for single-robot
solutions~\cite{Brambilla2013}. Typical examples include planetary
exploration~\cite{Goldsmith1999}, deep underground
mining~\cite{Rubio2012}, ocean restoration, and
agriculture.

A common aspect in these scenarios is the necessity to maintain a
coherent state across the swarm. Many basic coordination problems can
be solved assuming low-bandwidth, occasional communication or even no
communication. However, global connectivity is an asset when
information must be exchanged in a timely manner, either to optimize a
global performance function, or to aggregate data in a sink. Task
allocation scenarios with stringent space and time constraints, such
as warehouse organization and search-and-rescue
operations~\cite{stormont2005} are prime examples of this category of
problems. In these scenarios, it is desirable for the robot network to
allow both short-range and long-range information exchange.

In this paper, we tackle the problem of deploying a robot network in a
decentralized fashion, under the constraint that long-range
information exchange must be possible at any time during a mission.
We assume that the robots must reach a number of distant
locations. While navigating to these locations, the robots must spread
without splitting the network topology in disconnected components. The
robots must achieve a final configuration in which data can flow
between any two target locations, using the robots as relays.

It is important to notice that it is not required for all of the
robots to take part in the final topology. Rather, it is desirable
that as few robots as possible are engaged in connectivity
maintenance, as this would free any extra robot for others tasks or to
act as occasional replacement for damaged robot in the topology. In
contrast, the robots that are part of the final topology must form a
persistent communication backbone that can be used by any robot when
necessary.

This aspect sets apart our work from existing research on connectivity
maintenance, which generally requires \emph{all} robots to be part of
the connected topology. The literature on this topic can be broadly
divided in two classes: algorithms in which the robots must attain a
final, static structure to maximize coverage~\cite{Aragues2014}, and
algorithms in which global connectivity is enforced while navigating
to a specific location as a single unit
(flocking)~\cite{Nestmeyer2017}. Our work, in contrast, aims to create
a dynamic, decentralized communication infrastructure that connects
specific locations and uses as few robots as possible.

Our approach assumes that the robots are initially deployed in a
compact, connected cluster. The robots then form a logical \emph{tree}
over the physical network topology. By growing the tree over time, the
distribution of the robots progressively and dynamically extends to
reach the target locations. The final configuration is a star-like
topology, in which data can flow between any two target
locations.

The main contributions of this work are:
\begin{enumerate}
\item The formalization of two algorithms to form and grow logical
  tree topologies that connect multiple target locations;
\item A comparative study of the algorithms, based on extensive
  physics-based simulations;
\item The validation of our findings through a large set of real-robot
  experiments.
\end{enumerate}

The rest of this paper is organized as follows. In \sect{probstat} we
formalize the problem statement. In \sect{approach} we present our
methodology. In \sect{evaluation} we report an evaluation of the
algorithms. In \sect{relatedwork} we discuss related work. The paper
is concluded in \sect{conclusions}.


\section{Problem Statement}
\label{sec:probstat}

\subsection{Robot Dynamics}
\label{sec:robotdynamics}

We consider $N$ robots with linear discrete dynamics
$$
x_i(t+1) = A x_i(t) + B u_i(t)
$$
where $x_i(t) \in \mathbb{R}^{2M}$ is the state of robot $i$ at time
$t$, $u_i(t) \in \mathbb{R}^{2M}$ is the control signal, and
$A,B \in \mathbb{R}^{2M \times 2M}$. The state $x_i(t)$ is defined as
$\left[p_i(t), v_i(t)\right]$, where $p_i(t) \in \mathbb{R}^M$
designates the position of robot $i$ and $v_i(t) \in \mathbb{R}^M$ its
velocity. State and controls are subject to the convex constraints
$$
\forall t \ge 0 \quad x_i(t) \in \mathcal{X}_i \quad u_i(t) \in \mathcal{U}_i.
$$
In this work we focus on 2-dimensional navigation ($M = 2$).

\subsection{Robot Communication}
\label{sec:robotcommunication}

We assume that the robots are capable of \emph{situated
  communication}. This is a communication modality in which robots
broadcast data within a limited range $C$, and upon receiving data, a
robot is able to estimate the relative position of the data sender
with respect to its own local reference frame.

We define the \emph{communication graph}
$\mathcal{G}_C = (\mathcal{V}, \mathcal{E}_C)$, where $\mathcal{V}$ is
the set of robots $\left\{1, \dots, N\right\}$, and
$\mathcal{E}_C \subseteq \mathcal{V} \times \mathcal{V}$ is the set of edges
connecting the robots. An edge $(i,j)$ between two robots exists at
time $t$ if their distance is within their communication range $C$,
i.e.,  $\parallel p_i(t) - p_j(t) \parallel \le C$.

\begin{definition}[Graph connectivity]
  A graph is \emph{connected} is there exists a path between any two nodes.
\end{definition}

Graph connectivity can be verified through well-known concepts in
spectral graph theory. From the definition of the graph adjacency
matrix
$$
A_{ij} =
\begin{cases}
  1 & \text{if }(i,j) \in \mathcal{E}_C \\
  0 & \text{otherwise}
\end{cases}
$$
and of the graph degree matrix
$$
D_{ij} =
\begin{cases}
  \sum_k A_{ik} & \text{if }i = j\\
  0             & \text{otherwise}
\end{cases}
$$
we can derive the Laplacian matrix $L = D - A.$ The graph is connected
if and only if the second smallest eigenvalue of $L$ is greater than
0. For this reason, this eigenvalue is called \emph{algebraic
  connectivity} or \emph{Fiedler value}~\cite{Fiedler1973}. We will
employ algebraic connectivity as a performance measure in the
experiments of \sect{evaluation}.

\subsection{Objectives}
\label{sec:objectives}

The objective of this work can be stated as follows: we aim to create
a progressive deployment strategy that can reach an arbitrary number
of geographically distant tasks while satisfying connectivity
constraints. In particular, the final configuration of the network
topology must allow communication between any two target locations.


\section{Approach}
\label{sec:approach}

\subsection{Roles}
\label{sec:roles}

In both algorithms, we assume that the robots are initially deployed
in a fully connected cluster. Subsequently, the robots must form a
tree by dynamically assuming a specific role in the process.

In both tree-forming algorithms, the robots can have four possible
roles: \emph{root}, \emph{worker}, \emph{connector}, or
\emph{spare}. The \emph{root} robot corresponds to the tree
root, and at any time during the execution only one robot can assume
this role. The \emph{worker} robots are the tree leaves, and they
correspond to robots that must reach the target locations, forcing the
tree to grow progressively. The \emph{connector} robots dynamically
join the tree to support its growth, leaving the pool of available
\emph{spare} robots.

\subsection{High-Level Behavior Specification}
\label{sec:behaviorspec}

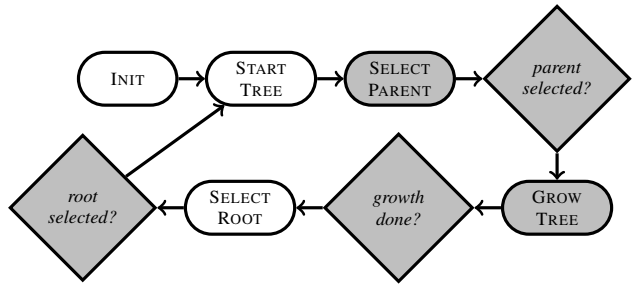
\begin{figure}[t]
  \centering
  \begin{tikzpicture}[
    state/.style={rounded rectangle,draw,very thick,minimum height=2em,text width=3em,align=center,font=\scriptsize\scshape},
    specstate/.style={state,fill=lightgray},
    barrier/.style={diamond,draw,fill=lightgray,very thick,minimum height=2em,text width=3em,align=center,font=\scriptsize\itshape},
    every on chain/.style={join=by {->,very thick}},
    start chain,
    node distance=1em
    ]
    \node(ii)[state,on chain]{Init};
    \node(it)[state,on chain]{Start Tree};
    \node(sp)[specstate,state,on chain]{Select Parent};
    \node(ps)[barrier,on chain]{parent selected?};
    \node(gt)[specstate,continue chain=going below,on chain]{Grow Tree};
    \node(tg)[barrier,continue chain=going left,on chain]{growth done?};
    \node(sr)[state,on chain]{Select Root};
    \node(rs)[barrier,on chain]{root selected?};
    \draw[->,very thick] (rs) -- (it);
  \end{tikzpicture}
  \caption{The high-level Finite State Machine that formalizes the
    individual robot behaviors in the two tree-formation
    algorithms. Rounded rectangles denote states, and diamonds denote
    \emph{barriers}, i.e., conditions that all robots must meet before
    proceeding to the next state. States filled in white are common
    among both algorithms; states and barriers filled in light gray
    differ across algorithms.}
  \label{fig:generalfsm}
\end{figure}

The algorithms can be formalized through a high-level state machine
that encodes the behavior of every robot, as depicted in
\fig{generalfsm}.

Every robot starts in state \textsc{Init}. We assume that a process
that assigns the role of \emph{worker} to the robots closest to the
targets has been already executed, through, e.g., a task allocation
algorithm or a gradient-based algorithm. In addition, a random robot
is assumed assigned the role of \emph{root}. The other robots are
initially \emph{spare}.

The \textsc{Start Tree} state is triggered by the root, which
propagates a signal throughout the robot network. This state signifies
that a new tree must be created. As the message propagates throughout
the network, the robots estimate their distance from the root. This is
possible because of situated communication---every robot can estimate
a relative vector to each of its immediate neighbors.

Robots receiving a ``start tree'' signal switch to \textsc{Select
  Parent}. In this state, each robot must identify a new parent to
attach to. The selection of a new parent aims to create the shortest
possible paths between the root robot and the \emph{worker} robots,
i.e., the leaf nodes in the tree. The specifics of this state are
different in the \emph{outwards} and \emph{inwards} algorithms, and
are explained in \sect{outwardsalg} and \sect{inwardsalg}. At the
end of this state, a robot is part of two trees---the one from the
previous iteration of the algorithm (excluding the very first
iteration), and a new one that reflects the new parent.

Once every robot has selected a new parent, the robots switch to the
\textsc{Grow Tree} state, in which the robots forget the tree from the
previous iteration and \emph{spare} robots are accepted to join an edge. 
The algorithms differ in the implementation of this state, and details
are reported in \sect{outwardsalg} and \sect{inwardsalg}.

Once the growth state is complete, the robots switch to the
\textsc{Select Root} state. As the tree grows, the initial choice of
the root robot (which is random) or an uneven distribution of target
locations might render the tree topology nonoptimal. By selecting a new
root, the swarm can balance the tree branches, thus fostering even
growth over time. The design of this state is illustrated in
\sect{rootselection}.

Finally, the new assigned root switches to state \textsc{Start Tree}
and broadcasts a new ``start tree'' signal.

In \fig{generalfsm}, certain state transitions are marked with
diamonds. These transitions, which we call \emph{barriers}, are
special in that they correspond to ``wait states'' in which the robots
must stay until a certain condition is verified for every robot. The
specific implementation of these conditions depends on the
algorithms. However, the general principle is that the root aggregates
the information necessary to evaluate a certain condition, and then
broadcasts a ``go'' signal throughout the tree. The ``go'' signal
triggers a state transition in the robots that receive it.

\subsection{Selection of a New Root}
\label{sec:rootselection}
The purpose selecting a new root is to balance the tree, which fosters
better growth and compensates for an uneven distribution of target
locations. In addition, balancing the tree has positive effects on the
scalability of our algorithms. Every state in our algorithms involves
some form of diffusion/aggregation process across the tree, with a
time complexity that is linear with the depth of the tree. By
balancing the tree, we also shorten its depth, thus lowering the time
for diffusion/aggregation processes to complete.

These considerations suggest that the best location for the root is as
close as possible to the centroid of the distribution of robots. The
selection of a new root occurs at the end of a tree configuration
loop, but the data upon which the process depends is collected in
state \textsc{Select Parent}, when the robots select a new parent.

\begin{figure}[t]
  \centering
  \includegraphics[height=5cm]{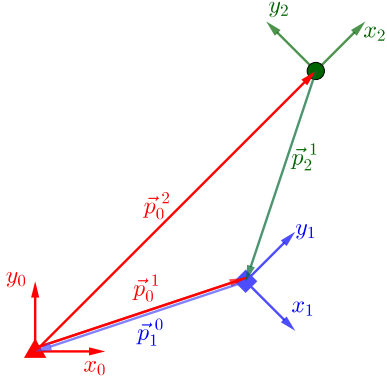}
  \caption{The red triangle represents robot 0 with the root reference
    frame. The blue square represents robot 1, which is a child of
    robot 0 and a parent of robot 2, in turn represented by the green
    circle.}
  \label{fig:centroid}
\end{figure}

The algorithm provides an estimate of the centroid in the root
reference frame by adding up each robot contribution from the leaves
to the root. The algorithm is formalized in \alg{centroid}. An
intuitive explanation of this algorithm proceeds as follows. Since
each robot only knows its relative position to other robots, it must
send to its parent an accumulation vector $ \mathbf{q}_i$ which
aggregates its contributions and that of all its descendants in the
tree, according to its own reference frame. \fig{centroid} reports an
example with three robots, where robot 0 is the root, robot 2 is a
\emph{worker}, and robot 1 is a \emph{connector}.


To perform the final calculation of the centroid, \alg{centroid} needs
the number of robots in the swarm. A tree-based distributed algorithm
to count the number of robots currently committed in the tree is
reported in \alg{count}. This algorithm requires the robots to
aggregate a partial count, denoted with $c_i$, from the tree leaves to
the root.

In our implementation, both \alg{centroid} and \alg{count} are
executed in parallel in state \textsc{Select Parent}. In
\textsc{select root}, the current root compares its position and the
position of its neighbors to the centroid estimate (all are expressed
in its reference frame). If the current root is the closest to the
centroid, it remains the root and restarts a new tree loop. Otherwise,
it designates a new root and sends the centroid vector and the angle
to the new root. When the new root receives this message, it sends an
acknowledgement message to the old root, and then it expresses the
centroid in its own reference frame. The process is repeated until the
root is the closest robot to the centroid estimate.

\begin{algorithm} [t]
  \begin{algorithmic}[1]
    \State $\mathbf{a}_i = 0$
    \ForAll{child $j$}
    \State $\mathbf{q}_j^i$ = express $\mathbf{q}_j$ in $i$'s reference frame
    \State $\mathbf{a}_i = \mathbf{a}_i + \mathbf{q}_j^i$
    \EndFor
    \If{robot $i$ has a parent}
      \State $\mathbf{q}_i =  \mathbf{a}_i - (\underbrace{c_i - d_i}_{\text{nb descendants}} + 1) \cdot \mathbf{p}^{\text{parent}}_i$
    \EndIf
    \If{robot $i$ is the root}
      \State $\mathbf{q}_i =  \mathbf{a}_i / \underbrace{c_i}_{\text{robot count}}$
    \EndIf
  \end{algorithmic}
  \caption{Distributed centroid estimation algorithm executed by robot
    $i$: $\mathbf{a}_i$ denotes an accumulator value; $\mathbf{q}_i$
    denotes the contribution of robot $i$ to the estimation algorithm;
    $c_i$ and $d_i$ denote the number of robots in the swarm estimated
    by robot $i$ and the tree depth of robot $i$, respectively; and
    $\mathbf{p}^{\text{parent}}_i$ is the vector from robot $i$ to its
    parent.}
  \label{alg:centroid}
\end{algorithm}

\begin{algorithm} [t]
  \begin{algorithmic}[1]
    \Switch{number of children}
      \Case{0}
        \State \textbf{return} $d_i$
      \EndCase
      \Case{1}
        \State \textbf{return} $c_{\text{child}}$
      \EndCase
      \Default
        \State \textbf{return} $\sum_{\text{neighbors }j} (c_j - d_i) + d_i$
      \EndDefault
    \EndSwitch
  \end{algorithmic}
  \caption{Tree-based count algorithm for robot $i$. The depth of
    robot $i$ in the tree is denoted as $d_i$. The depth of the tree
    root is set to 1. The count calculated by robot $j$ is denoted as
    $c_j$.}
  \label{alg:count}
\end{algorithm}

\subsection{The Outwards Algorithm}
\label{sec:outwardsalg}
The intuition behind the outwards algorithm is to build a logical
spanning tree over the entire robot network. The process starts at the
root, and robots join the tree progressively.

In state \textsc{Select Parent}, robot $i$ considers its neighbors as
potential candidates. Viable candidates are non-\emph{workers} already
in the tree and at a distance smaller than the communication
range. Among these, the robot selects the closest robot. The robot
commits to the tree and starts broadcasting its parent id, which
indicates to the parent robot that robot $i$ is a child and that $i$
is a \emph{connector}. Each \emph{connector} maintains its list of
children and checks for obstructions of line-of-sight with respect to
its parent. If a robot can not receive data from its selected parent,
it selects another parent and updates its data.

In state \textsc{Grow Tree}, the robots undergo two main phases:
first, they discard the information about the old tree; second, they
prune tree branches that contain no \emph{workers}. To establish
whether a branch contains a \emph{worker}, when a \emph{worker}
selects a parent (state \textsc{Select Parent}), the latter propagates
this information upstream towards the root.

\begin{figure}[t]
  \centering
  \includegraphics[height=5cm]{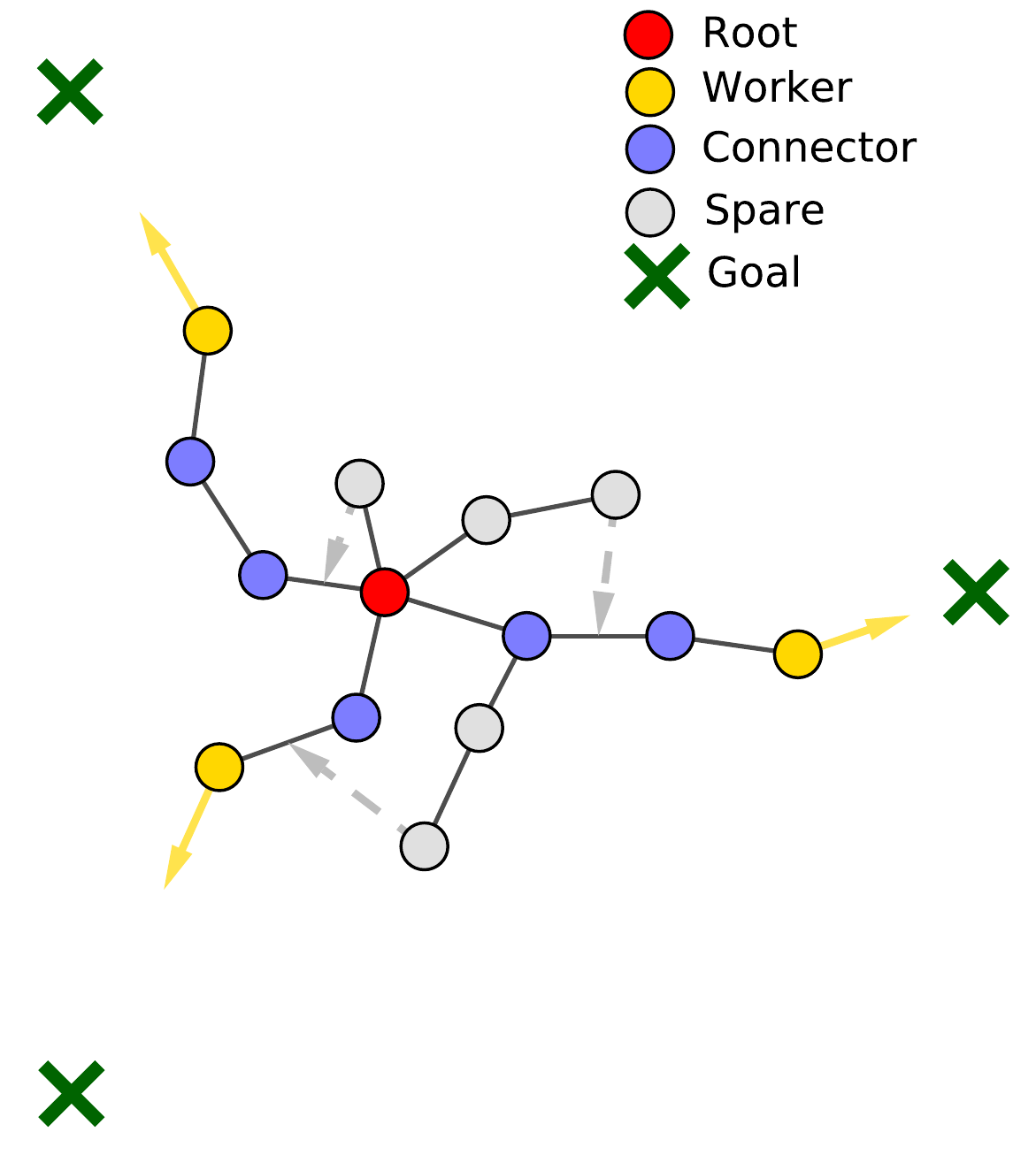}
  \caption{Spare management in the \emph{outwards} algorithm. The
    useful tree edges (blue nodes) are extended by pruning useless
    tree branches (grey nodes).}
  \label{fig:outwardsspare}
\end{figure}
The branches not containing a \emph{worker} are considered ``useless''
and the robots that are part of them take the \emph{spare} role. To
disband a useless branch, \emph{spare} robots leave it starting from
the leaves. The leaves curl the branch back towards the root, and upon
entering in contact with another branch might decide to join it. The
logic for spares to join a branch is explained in
\sect{sparemanegement}

\subsection{The Inwards Algorithm}
\label{sec:inwardsalg}

The intuition behind the \emph{inwards} algorithm is that the robots
join the tree starting from the workers towards the root. Growth is
therefore directed, and the final topology is a \emph{sparse} tree, in
that only a subset of the robots takes part in it. The \emph{spare}
robots, in contrast to the \emph{outwards} algorithm, do not form
branches; rather, they disperse along the tree and select a robot to
use as reference.

\begin{figure}[t]
  \centering
  \includegraphics[height=5cm]{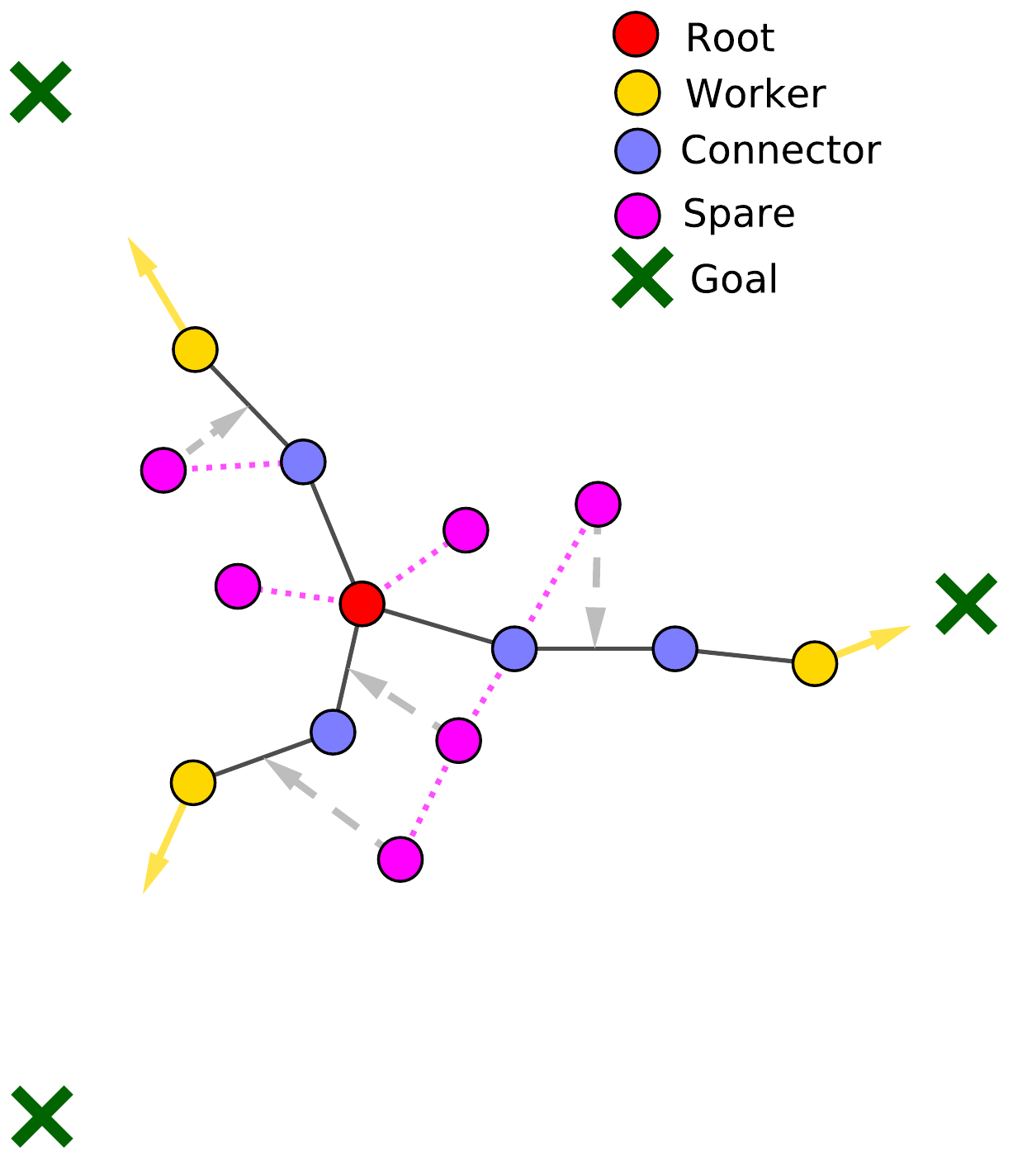}
  \caption{Spare management in the \emph{inwards} algorithm. Useful
    tree edges (blue nodes) are extended by adding \emph{spare} robots
    (purple nodes)}
  \label{fig:outwardsspare}
\end{figure}
In state \textsc{Select Parent}, viable candidates for parent
selection are non-workers in the tree or robots not in the tree which
are at a distance smaller than the communication range $C$. Among
these, a robot selects a neighbor with the smallest distance to the
root. When the robot $i$ commits to the tree, it broadcasts its parent
id, which indicates to the parent robot that robot $i$ is a child and
that $i$ is a \emph{connector}. In the \emph{inwards} algorithm, by
definition, all branches are useful because they all terminate with a
\emph{worker} as leaf node.

In state \textsc{Grow Tree}, \emph{spare} robots attempt to join a
branch. The logic for branch joining is the same as in the
\emph{outwards} algorithm, and it is explained in
\sect{sparemanegement}.

\subsection{Spare Management}
\label{sec:sparemanegement}
\begin{figure}[t]
  \centering
  \includegraphics[width=.48\textwidth]{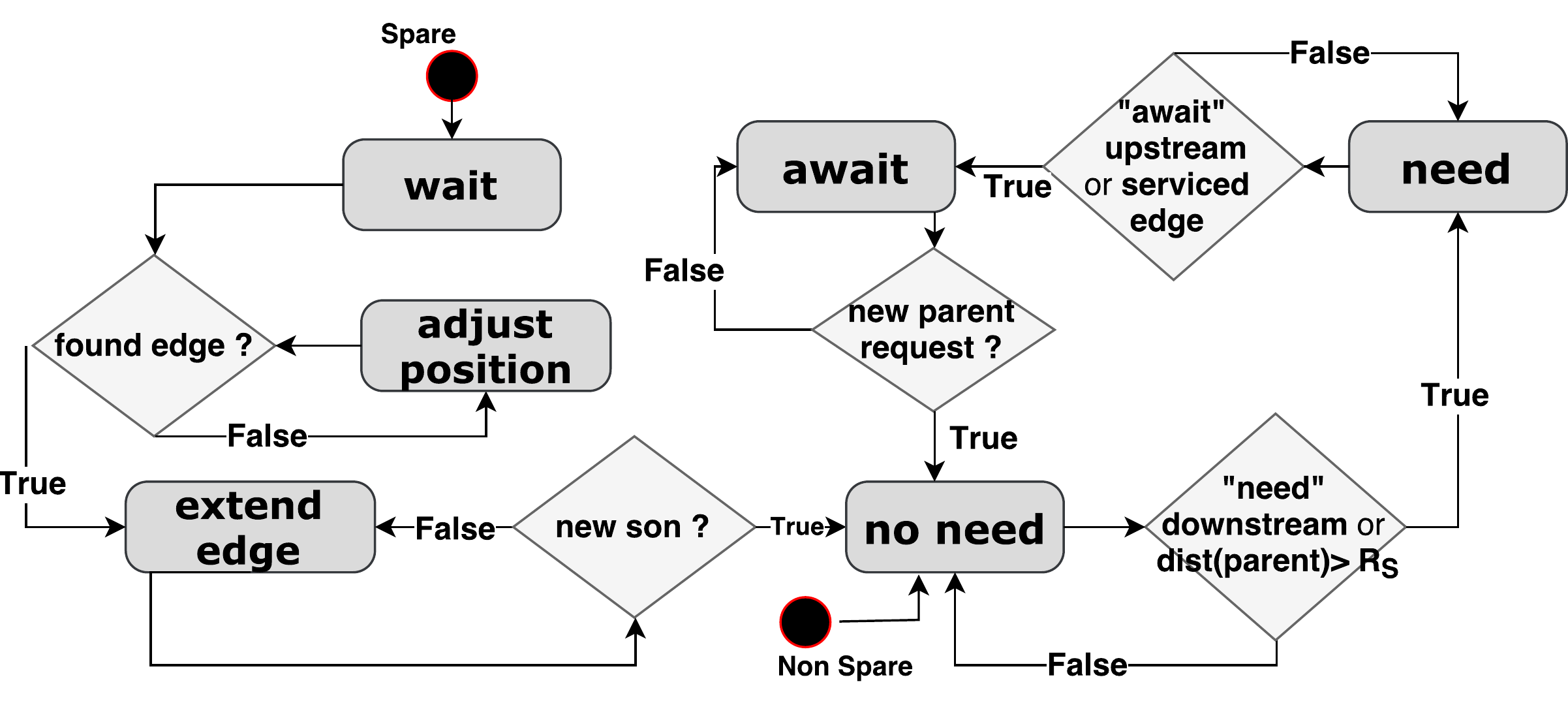}
  \caption{Interaction between \emph{spare} and \emph{non-spare}
    robots.}
  \label{fig:sparerobot}
\end{figure}

The state machine diagram in \fig{sparerobot} describes the part of
the \textsc{Grow Tree} state that concerns the interaction between
\emph{spare} robots and non-\emph{spare} robots (i.e.,
\emph{connectors}, \emph{workers}, and \emph{root}).

Non-spare robots enter the \textsc{no need} state when they have no
need for a spare robot. They exit this state either if their distance
to their parent becomes smaller than the \emph{safe communication
  range} $S$, or if at least one of their children's state is the
\textsc{need} state.  In the \textsc{need} state, each robot
continuously checks if it is in an edge selected by a \emph{spare}
robot, or if their parent is in the \textsc{await} state. If one of
these conditions is fulfilled, the robot transitions to the
\textsc{await} state.  In the \textsc{await} state, the robot is
waiting the insertion of a spare robot either in one of its edges or
upstream in the tree.

\emph{Spare} robots enter the \textsc{wait} state and look for an edge
to extend. They transition to the \textsc{extend edge} state or the
\textsc{adjust position} state after performing a search for edges in
need among their neighbors.  In the \textsc{adjust position} state,
spare robots rotate around their parent if they are within the safe
radius or move towards their parent in a straight line otherwise.  In
the \textsc{extend edge} state, spare robots head for the middle of
the edge to be extended.

\subsection{Robot Motion}
\label{sec:robotmotion}

The integrity of the tree over time is ensured by constraining the
robots' motion. We enforce the constraints by expressing the robot
motion as a sum of virtual potential forces (we omit time dependency
for brevity of notation):
$$
u_i =
\begin{cases}
  u_i^{\text{tree,old}} + u_i^{\text{tree,new}} \\ \quad\quad\quad + f_i(d_i^{\text{parent}}) (u_i^{\text{target}} + u_i^{\text{avoid}}) & \text{if }d_{i,j} \le E\\
  \mathbf{p}_i^{\text{parent}} & \text{otherwise} \\
\end{cases}
$$
where $d_{i,j} = \parallel p_i - p_j \parallel$, $E < C$ is the
\emph{emergency} range beyond which a robot is dangerously distant
from its parent, and
\begin{itemize}
\item $u_i^{\text{tree,old}}$ and $u_i^{\text{tree,new}}$ indicate the
  interaction law between robots $(i,j)$ in a parent-child
  relationship, in either the old or the new tree. We use the control
  law
  $$
  u_i^{\text{tree}} =
  \frac{\epsilon}{d_{i,j}}
  \left(
    \left(
      \frac{\delta}{d_{i,j}}
    \right)^2
    -
    \left(
      \frac{\delta}{d_{i,j}}
    \right)^4
  \right)
  $$
  where $\delta = E$ and
  $\epsilon$ are parameters to set at design time.
\item $u_i^{\text{target}}$ is a control law that attracts a robot to a
  target, promoting tree growth. For workers, this is a force that
  points the assigned target location $l_i$ and calculated with
  $$
  u_i^{\text{target}} = \tau \frac{l_i - p_i}{\parallel l_i - p_i \parallel}
  $$
  where $\tau$ is a design parameter. Workers propagate to their
  parents the calculated $u_i^{\text{target}}$, and connectors apply it in turn.
\item $u_i^{\text{avoid}}$ is a repulsive force for obstacle avoidance
  between neighbors not in a parent-child relationship.
\item $f_i(d_i^{\text{parent}})$ is a function defined as follows:
  $$
  f_i(d_i^{\text{parent}}) =
  \begin{cases}
    1 & \text{if }d_i^{\text{parent}} \le S\\
    0 & \text{otherwise}
  \end{cases}
  $$
  where $d_i^{\text{parent}}$ is the distance between a robot and its
  parent and $S < E$ is the \emph{safe communication range}. Through
  this function, a robot can ignore navigation to target and obstacle
  avoidance to perform emergency maneuvers when the distance to its
  parent becomes unsafe.
\end{itemize}


\section{Evaluation}
\label{sec:evaluation}

\subsection{Parameter Setting}
\label{sec:paramopt}

\begin{table*}[t]
  \centering
  \caption{Optimized Design Parameters}
  \label{tab:paramopt}
  \begin{tabular}{|c|c|l|c|c|c|}
    \hline
    Type & Symbol & Meaning & Outwards & Inwards & Unit\\
    \hline
    \hline
    \multirow{5}{*}{Motion}
    & $S$ & Safe range between parent and child & 138.93 & 135.25581 & cm \\
    \cline{2-6}
    & $A$ & Non-parent-child avoidance range & 43.16 & 40.99 & cm\\
    \cline{2-6}
    & $\delta$ & Ideal distance between parent and child & 190 & 154.0841 & cm\\
    \cline{2-6}
    & $\epsilon$ & Factor gain in parent-child interaction & 10 & 10 & \\
    \cline{2-6}
    & $\tau$ & Magnitude of attraction to target & 0.49 & 0.2539 & \\
    \hline
    \multirow{2}{*}{Tree Growth}
    & $R$ & Reconfiguration period & 38.8 & 44.0 & sec\\
    \cline{2-6}
    & $I$ & Information liveness period & 1.2 & 0.5 & sec\\
    \cline{2-6}
    \hline
    \multirow{2}{*}{Uncommitted Management}
    & $E$ & Distance threshold for \emph{spare} recruitment & 132.09 & 132.1353 & cm\\
    \cline{2-6}
    & $J$ & Distance threshold to switch to \emph{connector} & 9.79 & 6.6395 & cm\\
    \cline{2-6}
    \hline
  \end{tabular}
\end{table*}

The dynamics and the performance of our algorithms depends on the
design parameters reported in \tab{paramopt}. To set their value, we
used a genetic algorithm. We ran multiple instances of the
optimization process for both \emph{inwards} and \emph{outwards}, and
\tab{paramopt} reports the best values we found.

Every instance of the optimization was executed for 100
generations. We set this number as a reasonable margin after observing
that, across instances, after about 50 generations the optimization
process would find a plateau beyond which no improvement was found.

Every generation consisted of trials in which 9 Khepera IV
robots\footnote{\url{https://www.k-team.com/mobile-robotics-products/khepera-iv}}
were placed in the arena in a tight cluster. We configured two types
of trials:
\begin{itemize}
\item 2 target locations on a circle with a radius of \unit[2.3]{m} at
  $\unit[180]{^\circ}$ from each other;
\item 3 targets on a circle with a radius of \unit[1.6]{m} at
  $\unit[120]{^\circ}$ from each other.
\end{itemize}

We ran the trials in the ARGoS multi-robot
simulator~\cite{Pinciroli2012}, and maximized a two-step performance
function. The first step (performance 0 to 1) promoted connectivity
maintenance by penalizing the time spent with disconnected robots; the
second step (performance 1 to 2) was activated when no disconnections
occurred, and higher values corresponded to lower times to reach the
targets.

\subsection{Simulated Experiments}
\label{sec:simexperiments}

We tested the performance of the algorithms by varying three
parameters: the target radius, the redundancy factor, and number of
targets. We placed multiple targets on a circle with equal angles
between each other. The \emph{target radius} is the radius of the
circle. We chose radii of 3, 6 and 9 meters corresponding to small,
medium and large scales. The \emph{redundancy factor} is the factor by
which we multiply the minimum required number of robots needed to
reach all the targets given our communication range. We tested the
values of 2, 3 and 4 for this parameter. The \emph{number of targets}
was 2, 3, and 4. The largest configuration we considered involved 94
robots. Each scenario was executed with 50 different random seeds. We
ran all the experiments for both algorithms with and without
activating line-of-sight obstructions in the communication models of
ARGoS, to test the effect of this aspect.

\subsubsection{Simulation Time}
\begin{figure*}[t]
    \centering
    \includegraphics[width=0.3\textwidth]{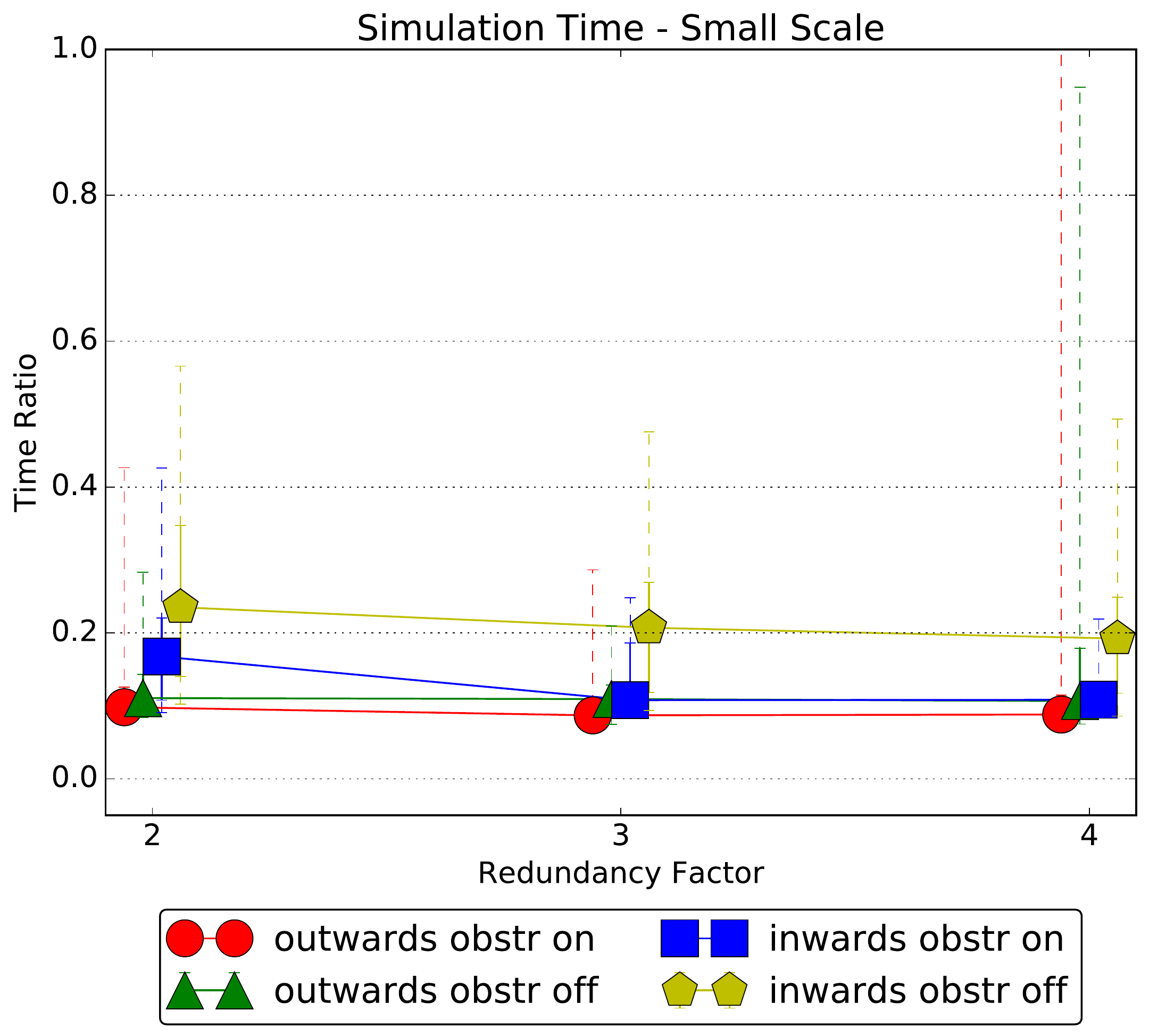}
    \includegraphics[width=0.3\textwidth]{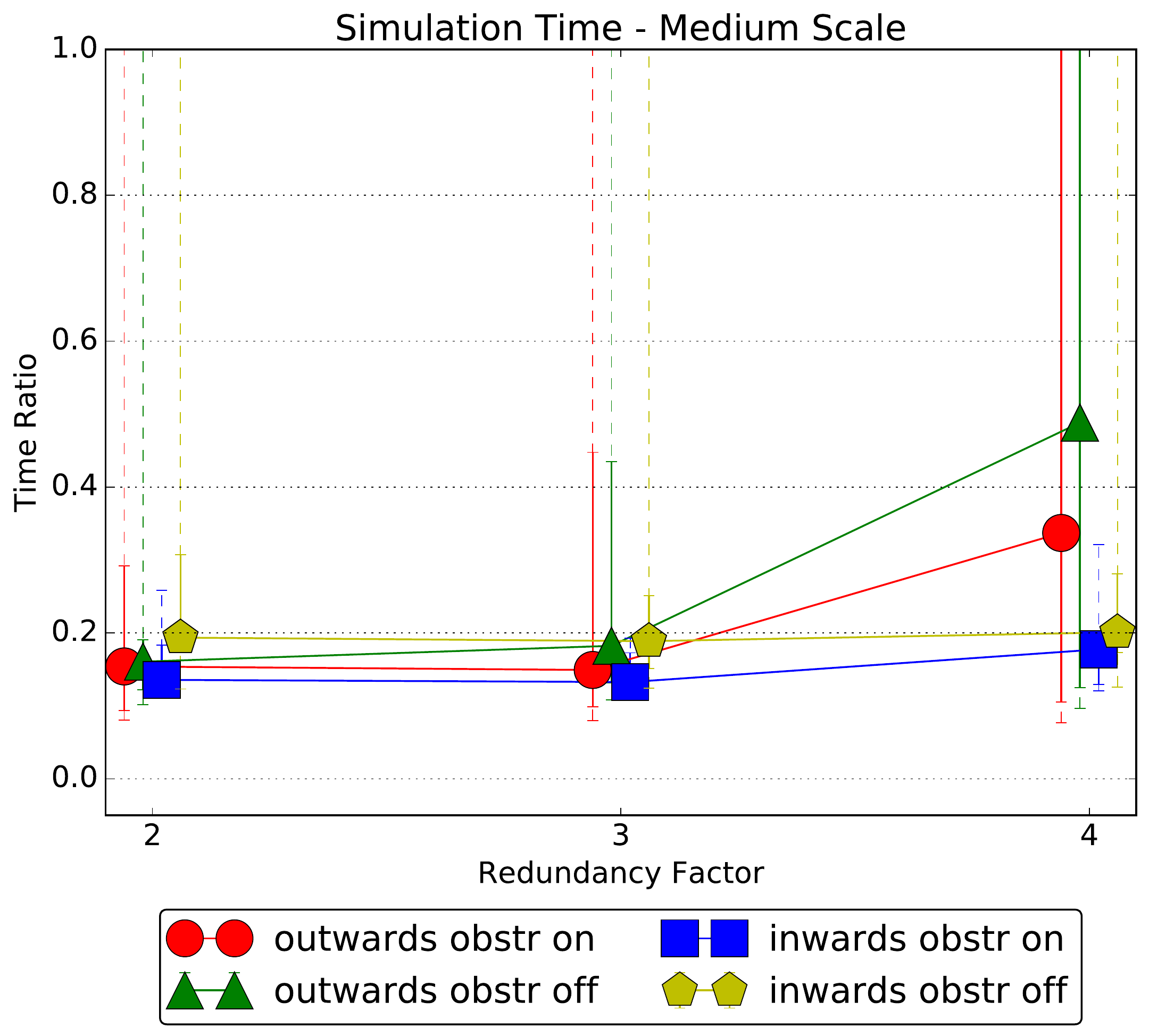}
    \includegraphics[width=0.3\textwidth]{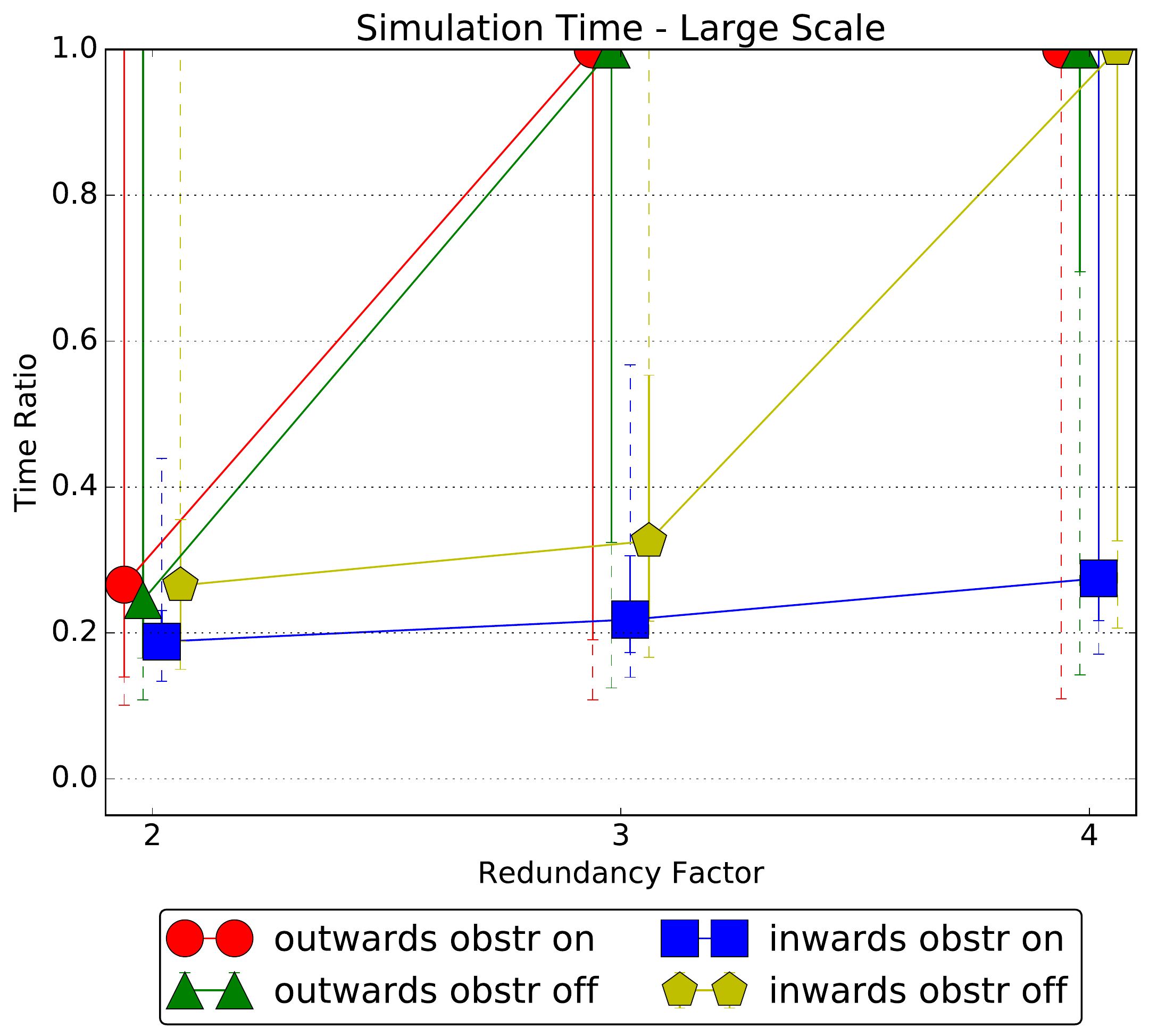}
    \caption{Assessment of mission completion time in simulation.}
    \label{fig:sim_time}
\end{figure*}

We studied the time performance of both algorithms, and declared an
experiment finished when all workers reach their targets. To compare
results across different scales, we normalized the mission duration by
the maximum allowed time. The maximum allowed time was computed by
considering the time for a robot to reach a target from the center of
the arena; this time was then multiplied by 10. The results are
reported in \fig{sim_time}. For small scales, the \emph{outwards}
algorithm outperforms the \emph{inwards} algorithm. However, as the
scale of the experiment is increased, the directed growth of the
\emph{inwards} algorithm is increasingly advantageous. In addition,
with the \emph{outwards} algorithm, some missions do not reach their
targets in the allotted time limits when higher redundancy factor is
employed. This is due to the increased interference that too many
useless branches create in robot navigation. This effect is not
prominent in the \emph{inwards} algorithm because the robots are added
to the tree only when it is necessary.

\subsubsection{Disconnected Time}
We studied the ability to maintain connectivity by considering the
following metrics:
\begin{inparaenum}[(i)]
\item The \emph{disconnected time ratio}, defined as the number of
  time steps (over the total experiment time) with at least a broken
  edge in the tree;
\item The \emph{Fiedler value time ratio}, defined as the number of
  time steps (over the total experiment time) with swarm-wide Fiedler
  value lower than $10^{-3}$.
\end{inparaenum}
\begin{figure*}[t]
  \centering
  \includegraphics[width=0.3\textwidth]{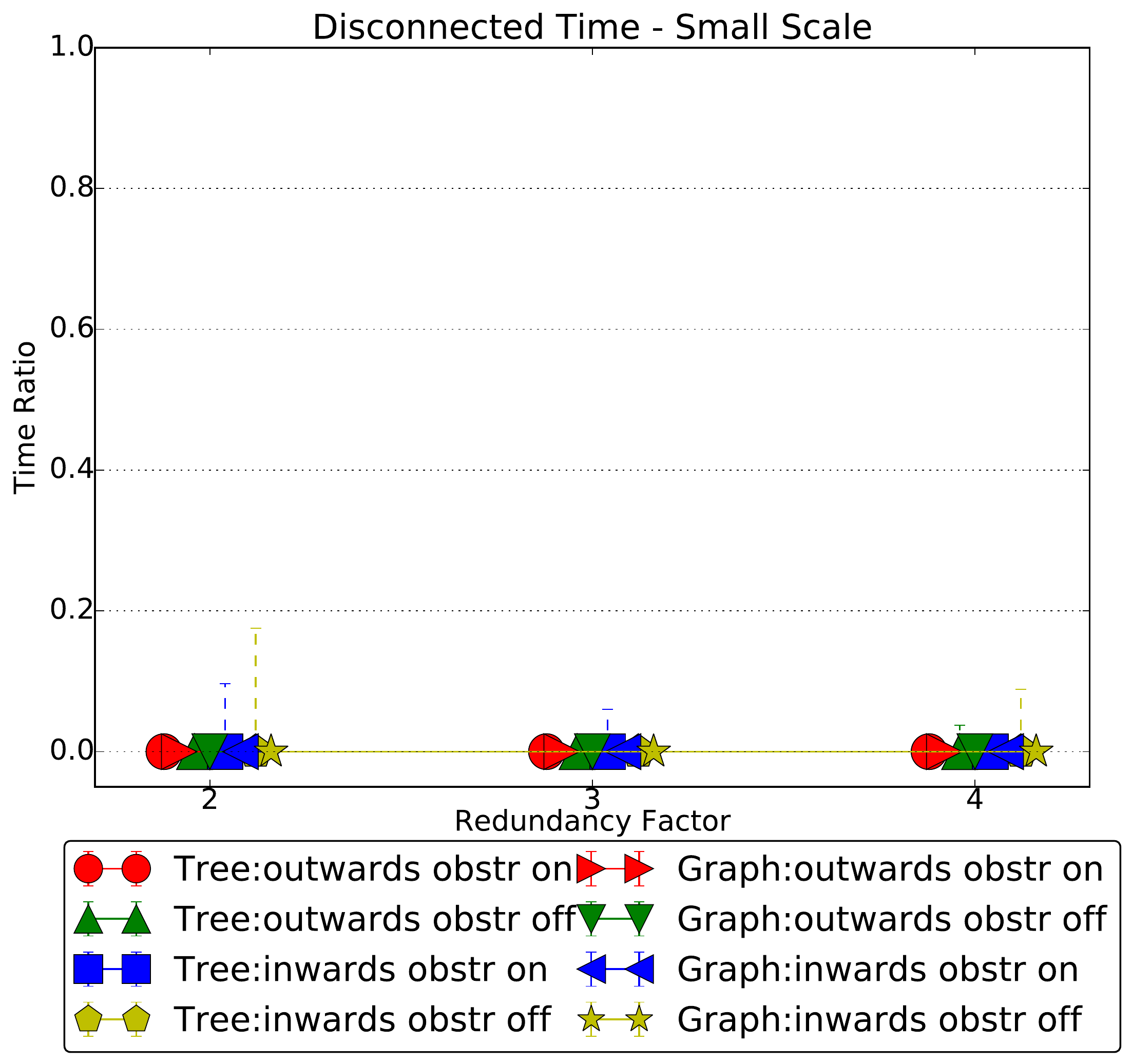}
  \includegraphics[width=0.3\textwidth]{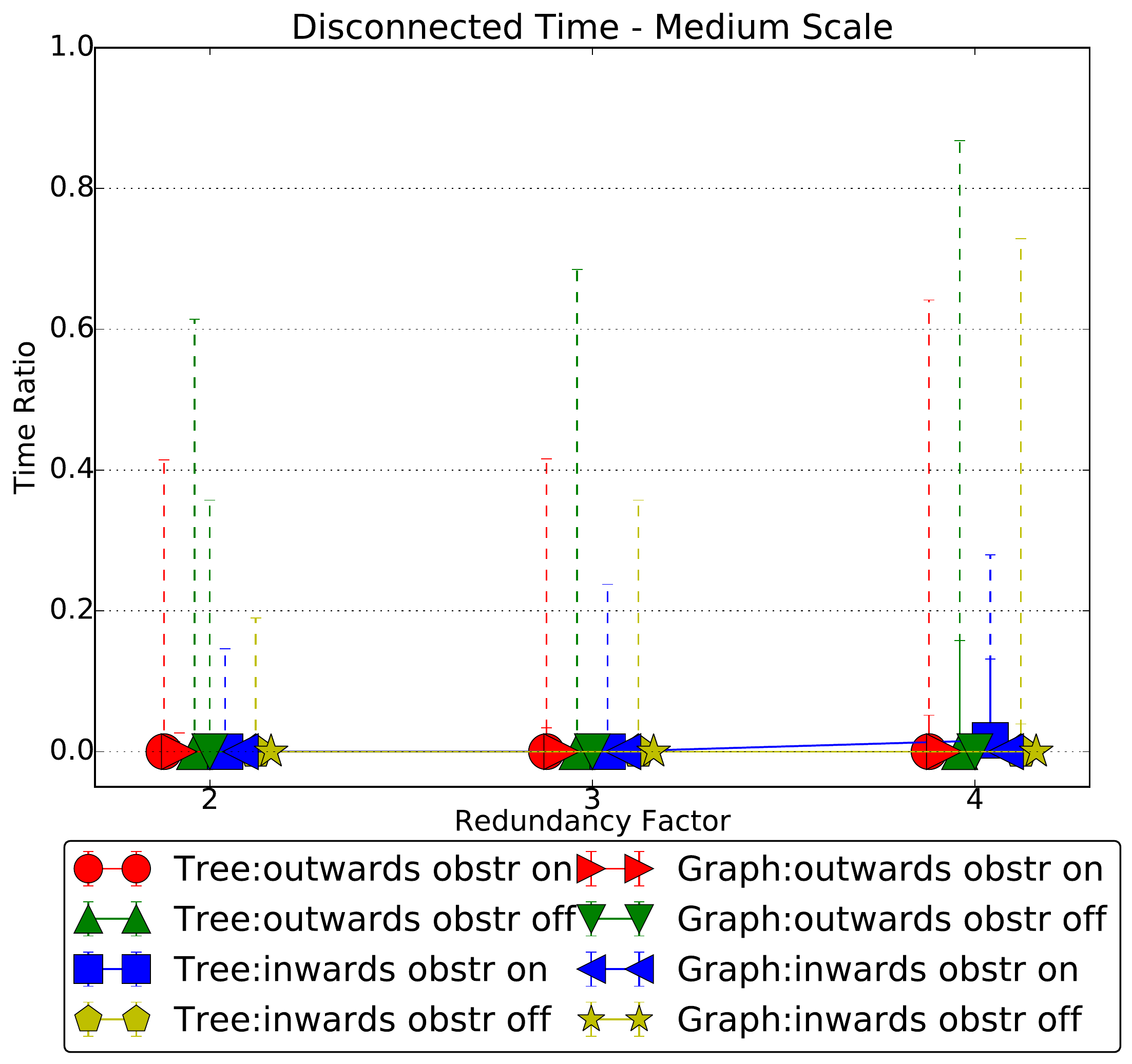}
  \includegraphics[width=0.3\textwidth]{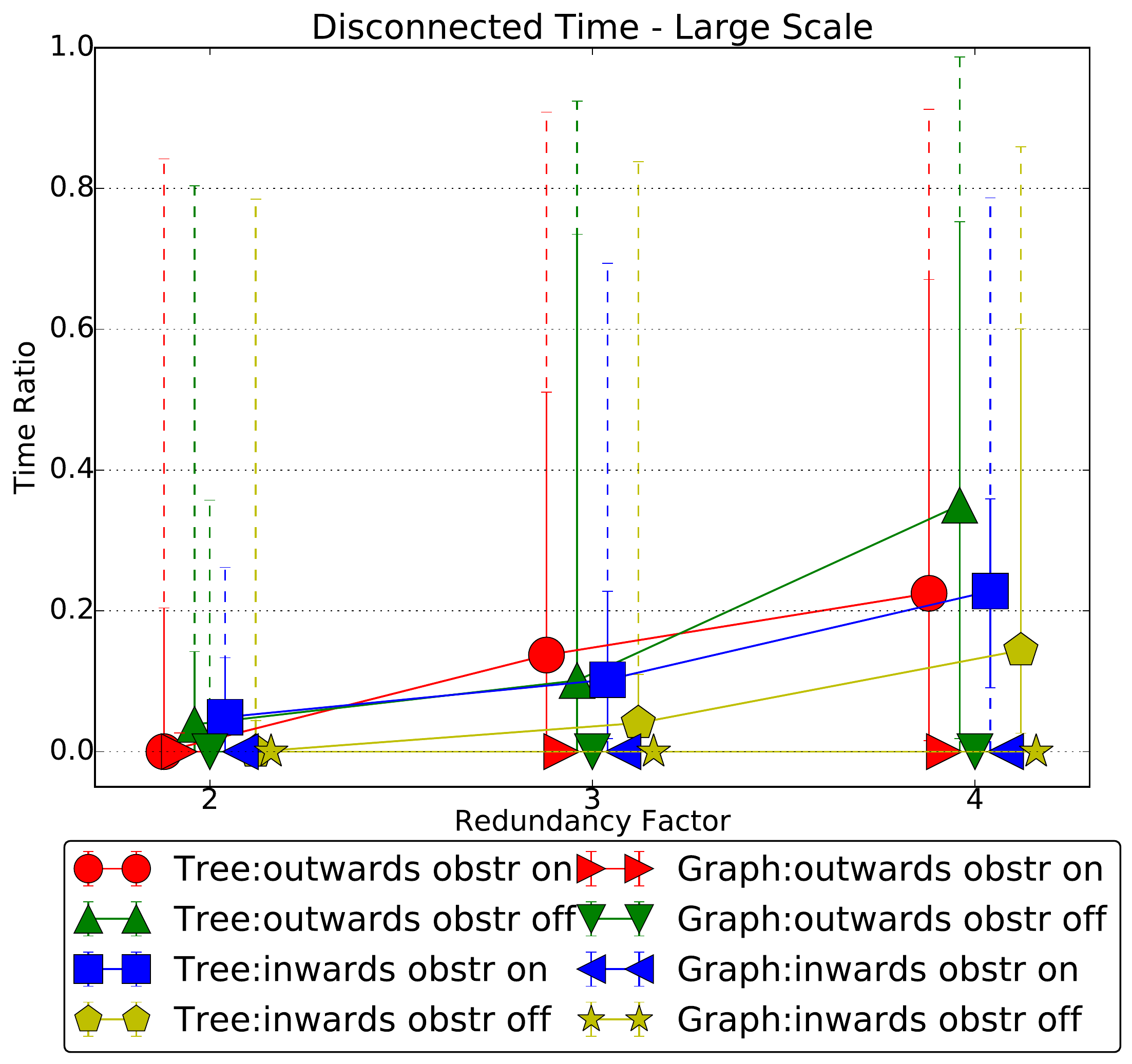}
  \caption{Assessment of connectivity loss.}
  \label{fig:disc_time}
\end{figure*}
The results are reported in \fig{disc_time}. In small-scale scenarios,
in only two experiments out of 50 have positive disconnected time, and
the global communication graph always stays connected. In medium-scale
scenarios, larger numbers of redundant robots cause occasional
line-of-sight obstructions that delay messages exchanges, but
connectivity is generally maintained throughout the duration of the
experiment. In large-scale scenarios, the disruptive effect of a large
number of redundant robots is prominent for both algorithms. With
fewer robots, the inwards algorithm is capable of maintaining global
connectivity in all of the experiments, despite occasional breaking of
tree edges (in less than 5\% of the experiments).













\subsection{Real-Robot Validation}
To validate the simulated results simulations, we tested our
algorithms with 9 Khepera IV robots. A Vicon motion capture system was
used to track the position and orientation of the robots throughout the
duration of the experiments, and to simulate situated communication.
We employed 2 experimental scenarios:
\begin{inparaenum}[(i)]
\item 2 targets on a circle with a radius of 2.3 meters at
  approximately 180 degrees from each other;
\item 3 targets on a circle with a radius of 1.6 meters at
  approximately 120 degrees from each other.
\end{inparaenum}
We rescaled the distance-related parameters in \tab{paramopt} to fit
the arena and accommodate for the small number of robots involved.  We
repeated these experiments 15 times for setup (i) and 10 times for
setup (ii) with robots starting from the same positions and
orientations, to allow for better comparison. We also performed the
same experiments in simulation, with the same initial positions.

\begin{figure}
    \centering
    \includegraphics[width=.3\textwidth]{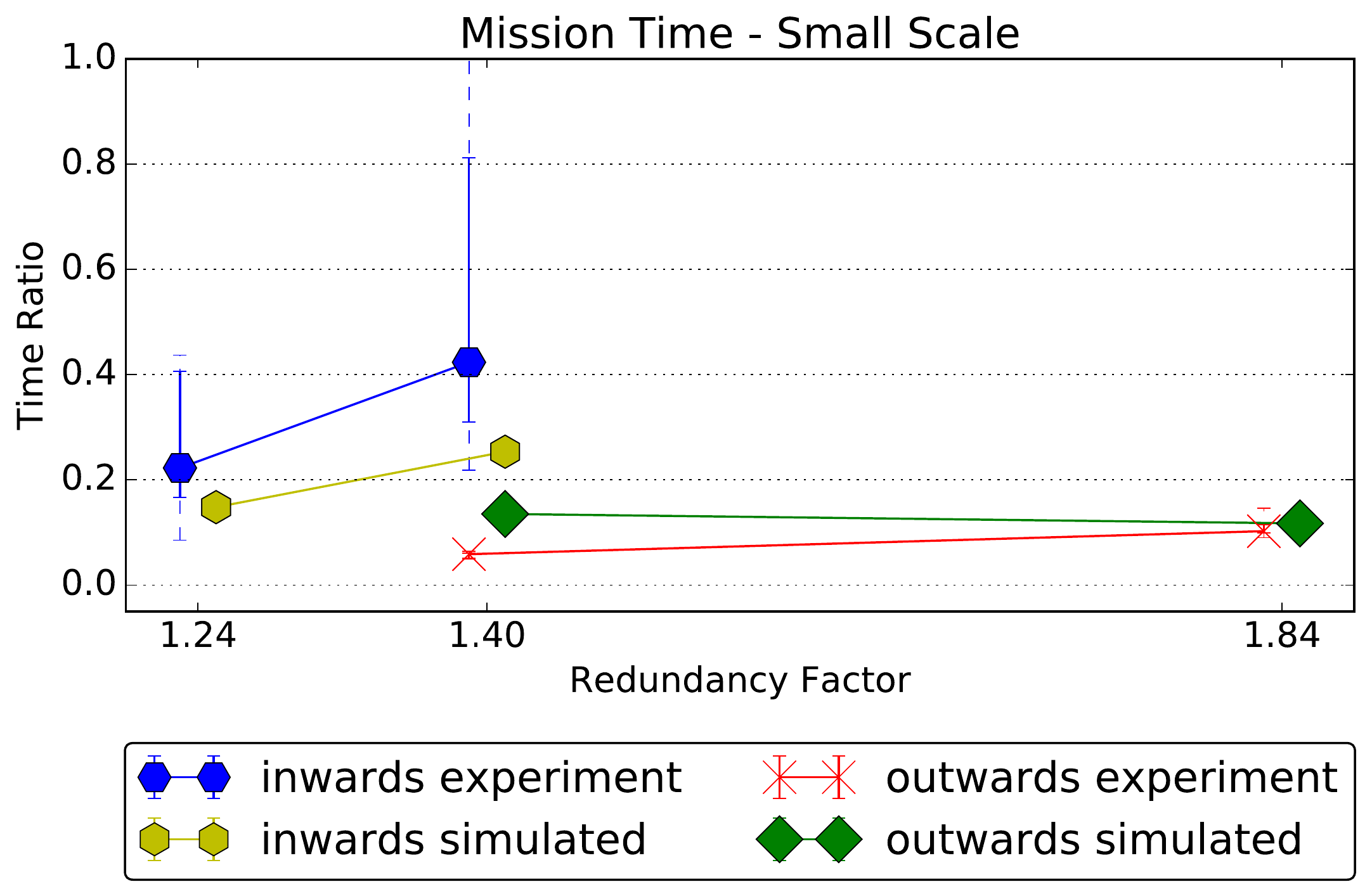}
    \includegraphics[width=.3\textwidth]{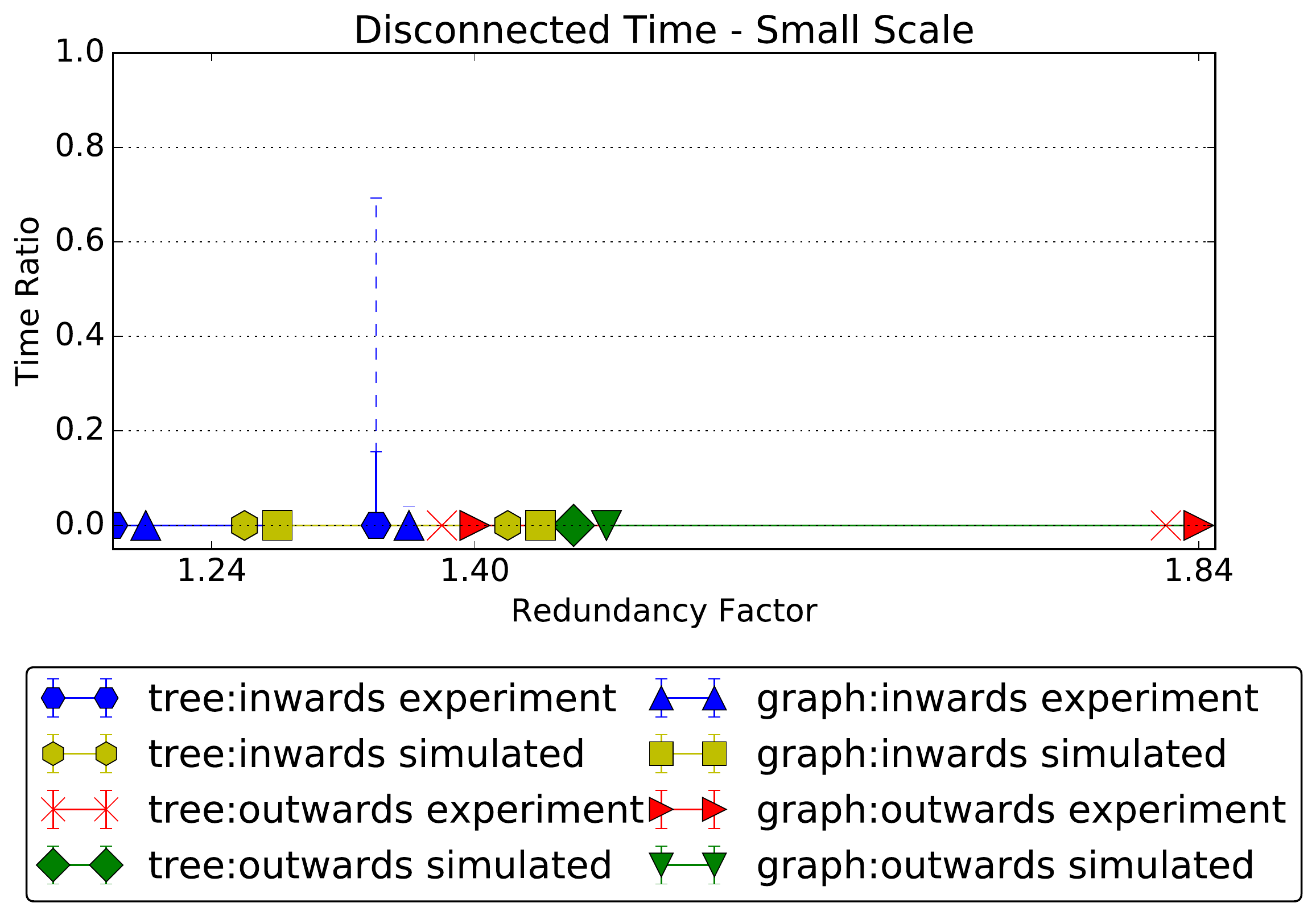}
    \caption{Results of real-robot evaluation.}
    \label{fig:realexperiment}
\end{figure}

\fig{realexperiment} shows that real-robot and simulated experiments
follow analogous trends. In particular, we verified that for
small-scale experiments with low redundancy factor (in these
experiments it was set 1) the \emph{outwards} algorithm has better
performance than the \emph{inwards} algorithm.

\section{Related Work}
\label{sec:relatedwork}

Extensive literature exists on methods for connectivity
preservation. Several recent works consist of motion control laws that
include an estimate of the Fielder value. Yang \emph{et
  al.}~\cite{Yang2010} introduced a decentralized algorithm to
estimate the Fiedler value and use it to maintain connectivity while
moving towards a target location. This algorithm was later refined by
Sabattini \emph{et al.}~\cite{Sabattini2011} and Williams \emph{et
  al.}~\cite{Williams2013}. Further extensions include inter-robot
collision avoidance~\cite{RobuffoGiordano2013} and multi-target
exploration~\cite{Nestmeyer2017}. The main advantage of this family of
approaches is that they allow navigation with arbitrary
topologies. However, accurate decentralized computation of the Fiedler
value is not easy in realistic settings in which messages might be
lost due to communication interference~\cite{DiLorenzo2013}. In
addition, computing the Fielder value in a decentralized manner
involves network-wide power iteration methods~\cite{Bertrand2013},
the slow convergence of which makes them suitable only for small teams of
robots~\cite{Sahai2012,Williams2013}. It should also be noted that all
of the above algorithms, with the exception
of~\cite{RobuffoGiordano2013}, have only been demonstrated in
simulated environments.

A second family of methods select a communication sub-graph and aim to
preserve its edges through some form of global consensus. Hsieh
\emph{et al.}~\cite{Hsieh2008} devised a reactive control law based on
radio signal and bandwidth estimation, in which links between robots
can be activated and deactivated as the topology changes over
time. Michael \emph{et al.}~\cite{Michael2009} employed distributed
consensus and auctions algorithms to establish which links to activate
and deactivate over time. Cornejo \emph{et
  al.}~\cite{Cornejo2009,Cornejo2012} proposed a distributed algorithm
for link selection in which the robots undergo a number of motion
rounds, during which the selected links must be preserved. Being based
on achieving global consensus before any topology modification can be
finalized, these algorithms are not scalable and work best when teams
involve a small number of robots.

A third class of connectivity-preserving algorithms assumes that a
certain structure is pre-existing. The dynamic structure is some form
of logical tree, dynamically built and updated over the physical links
of the robot network. Our work falls into this category. Krupke
\emph{et al.}~\cite{Krupke2015} employed a Steiner tree as a
pre-existing structure, and use spring-like virtual forces to balance
connectivity and cohesiveness while reaching distant targets. A number
of works, which constitute our main source of inspiration, utilized
minimum spanning trees as structures to preserve.  Aragues \emph{et
  al.}~\cite{Aragues2014} focused on a distributed coverage strategy
with connectivity constraints, and proposed a method based on
maintaining a network-wide minimum spanning tree. Analogously,
Soleymani et al.~\cite{Soleymani2015} proposed a distributed approach
that constructs and preserves a network-wide minimum spanning tree,
allowing for tree switching. Schuresko \emph{et
  al.}~\cite{Schuresko2012} studied a theoretical approach for
distributed and robust switching between minimum spanning trees.  All
these works were only demonstrated in numerical simulations. The main
advantage of these methods is the ease and speed with which spanning
trees can be built and updated in a distributed manner. However, as
discussed in this paper, spanning trees do not scale well with the
number of robots involved.

\section{Conclusions}
\label{sec:conclusions}

In this paper, we presented two algorithms to construct a long-range
communication backbone that connects multiple distant target
locations. The algorithms are decentralized and based on the idea of
constructing a logical tree over the set of physical network links.

We performed an extensive large set of experiments, both in simulation
and with real robots, to assess the performance of the algorithms
according to various experimental conditions. Our results show that,
in small-scale scenarios, \emph{outwards} tree growth, corresponding
to spanning tree formation, is a viable approach. However, as the
scale of the environment and the number of robots involved increase, a
more directed, \emph{inwards} growth from target locations towards the
tree root, is a preferable approach.

Our results also show that, as the number of unnecessary robots
increases, the benefit of redundancy is voided by the increased
physical interference in navigation. While a better spare robot
strategy could diminish this phenomenon, our results suggest that a
more progressive approach to deployment might be a better idea.

Nonetheless, the presence of a reasonable number of spare robots
offers the opportunity to tackle the problem of maintaining
\emph{persistent} long-range global connectivity despite individual
limitations in the energy supply of individual robots. We plan to
consider this scenario in future research.

In addition, possible extensions of our work include the presence of
moving targets, rather than static ones, and the presence of obstacles
in the environment.


\bibliographystyle{IEEEtran}
\bibliography{biblio}

\addtolength{\textheight}{-12cm}   

\end{document}